# Corpus-based Method for Automatic Identification of Support Verbs for Nominalizations

Gregory Grefenstette
Rank Xerox Research Centre
38240 Meylan, France
grefen@xerox.fr

Simone Teufel
Universität Stuttgart
Institut für maschinelle Sprachverarbeitung
D 70174 Stuttgart 1
simone@ims.uni-stuttgart.de

## Abstract

Nominalization is a highly productive phenomena in most languages. The process of nominalization ejects a verb from its syntactic role into a nominal position. The original verb is often replaced by a semantically emptied support verb (e.g., *make a proposal*). The choice of a support verb for a given nominalization is unpredictable, causing a problem for language learners as well as for natural language processing systems. We present here a method of discovering support verbs from an untagged corpus via low-level syntactic processing and comparison of arguments attached to verbal forms and potential nominalized forms. The result of the process is a list of potential support verbs for the nominalized form of a given predicate.

## 1 Introduction

Nominalization, the transformation of a verbal phrase into a nominal form, is possible in most languages (Comrie and Thompson, 1990). Nominalizations are used for a variety of stylistic reasons: to avoid repetitions of a verb, to avoid awkward intransitive uses of transitive verbs, in technical descriptions where passive is commonly used, etc.

Though, as a result of nominalization, the original verb is ejected from its syntactic position, it often retains many of its thematic roles. The original agents and patients can reappear as genitival or adjectival modifiers of the nominalized predicate. In the syntactic place of the original verb can appear a semantically impoverished verb.

The semantically impoverished verb, often called a support verb, to be used with a nominalized predicate structure is unpredictable. Allerton (1982)[p. 76] writes:

> Perhaps the most serious problem for these structures is that there is no constant selection of the empty verb: sometimes we find *have*, sometimes *take*, sometimes *give*, and rarely *pay*; sometimes we have a choice between two or more empty verbs e.g. *have/take a look* ... We have little choice but to record such irregularities in the lexicon ...

For this reason, the collocational choice of a support verb for a given nominalization is a difficult problem for language learners, as well as for natural language processor implementation.

We present here a method of deriving probable support verbs for nominalized predicates from corpora using low-level syntactic analysis and simple frequency statistics over its results. This automatic procedure may be looked upon as an aid to lexicographers, as an independent extraction tool, or as a verification of lexical collocation information stored in a machine-based lexicon.

Since nominalized predicates can semantically drift over time to become concrete nouns having lost all their thematic role, our method first attempts to distinguish true nominalizations from concrete uses of the nominalized surface form. This is done by comparing approximations to the argument/adjunct structures of verbal predicates to those of candidate nominalized forms in a large corpus. For each selected nominalized form, syntactically supporting verb information is extracted from the corpus and then collated, providing the candidates for support verbs.

## 1.1 The Nominalization Cline

The phenomenon of nominalization in English happens when a verb is replaced by a noun construction using a gerundive or nominal form of the verb. The original subject and objects of the verb can reappear as Saxon or Norman genitives modifying the nominalized form.

Quirk et al. (Quirk et al., 1985) distinguish nominalizations between deverbal and verbal nouns. Examples of these are *advice* vs. *killing*. Deverbal nouns are defined as records of the action having taken place rather than as description of the action itself. This accounts for the contrast between *their arriving for a month* and *\*their arrival for a month*.

Deverbal nouns can be replaced by regular count nouns in any context, for example *painting* as a deverbal noun in *Brown's paintings of his daughter* can be replaced by *photograph* whereas this is not the case with the verbal noun *painting* in *The painting of Brown is as skillful as that of Gainsborough* which describes the action of painting itself (Quirk et al., 1985)[p.1291].

As the following evidence shows, the semantics of the verb and much of its syntactic structure can be retained by either of its nominalized forms:

> *She was surprised that the enemy destroyed the city.*
> *She was surprised by the enemy('s) destroying the city.*
> *She was surprised by the enemy's destruction of the city.*

The cline of nominalization can be seen in the morphological changes[1] that some predicate undergo as they move from an inflected verb (e.g. *destroyed*) to non-inflected verbal noun (*destroying*) to a deverbal noun (*destruction*).

In this article we shall consider only deverbal nouns[2] since these are the nominalizations involving the collocational phenomena of support verbs.

A remaining problem with the deverbal nouns is that the meaning of such nouns can become concretized over time, by a metonymic association. Compare the uses of *proposal* in :

> *He made his formal proposal to the full committee.*
> *He put the proposal in the drawer.*

The concrete uses of deverbal nouns are not involved in support verb constructions since they have lost the semantics of actions, and their attending thematic roles.

## 2 A very simple approach and its problems

Looking for support verbs for nominalizations might seem an easy problem at first, given that these support verbs are always the main verb for which the nominalization is the direct object. What is needed is a low-level parser that extracts verb-object relations from corpora. Given such a parser, one might be tempted to extract all the main verbs for a given nominalized form and consider the most frequent of these verbs the expected support verbs.

As will be seen below, this approach is too simple. The examples given above for *proposal* show that a given word form may be used with a meaning anywhere along the cline from true nominalization to concrete nouns. Counting verbs of which these concrete nouns are direct objects will create noise hiding the true support verbs.

Since real nominalizations are those that still have verbal character, i.e. they have retained the semantic roles from the verb, we will try to recognize true nominalized uses by comparing the most frequent argument/adjunct structures found in the corpus around verbal uses of a given predicate to those syntactic structures found around the candidate nominalized forms[3].

We will define true nominalizations as those which have a parallel syntactic structure to the original verb. This is also in keeping with the definition of nominalizations given in (Quirk et al., 1985)[p. 1288]:

> A noun phrase ...which has a systematic correspondence with a clause structure will be termed a

---

[1] The morphological processes involved in transforming verbs into nominalizations are described in (Quirk et al., 1985), Sections I.43 (conversion) and I.30 (suffixation). See also 17.52 for discussion of this cline.

[2] On a practical level, we will also accept as deverbal nouns those forms ending in *-ing* which are marked as nouns in our lexicon, e.g. *warning*.

[3] Since these argument/adjunct structures are difficult to recognize precisely without an elaborate parser incorporating semantic analysis, we decided to identify heuristically the structures fulfilling these roles, for example, taking the most frequently occurring prepositional phrases after verbal sequences as adjuncts or arguments.

**Nominalization.** The noun head of such a phrase is normally related morphologically to a verb, or to an adjective (i.e., a deverbal or deadjectival noun).

In this article we have been restricting the term nominalization to the noun heading the noun phrase.

## 3 Method

In this section we explain our method for extracting support verbs for nominalizations. We suppose that we are given a pair of words: a verb and its nominalized form. As explained in the previous section, we are interested in extracting only nominalized forms which have not become concrete nouns, and that this will be done by comparing syntactic structures attached to the verb and noun forms. In order to extract corpus evidence related to these phenomena, we proceed as follows:

1. We generate all the morphologically related forms of the word pair using a lexical transducer for English (Karttunen et al., 1992). This list of words will be used as corpus filter.

2. The lines of the corpus are tokenized (Grefenstette and Tapanainen, 1994), and only sentences containing one of the word forms in the filter are retained.

3. The corpus lines retained are part-of-speech tagged (Cutting et al., 1992). This allows us to divide the corpus evidence into verb evidence and noun evidence.

4. Using a robust surface parser (Grefenstette, 1994), we derive the local syntactic patterns involving the verbal form and the nominalized form.

5. Considering that nominalized forms retain some of the verbal characteristics of the underlying predicate, we want to extract the most common argument/adjunct structures found around verbal uses of the predicate. As an approximation, we extract here all the prepositional phrases found after the verb.

6. For nominal forms, we select only those uses which involve argument/adjunct structures similar to phrases extracted in the previous step. For these selected nominalized forms, we extract the verbs of which these forms are the direct

| *frequency* | |
|---|---|
| 458 | million |
| 438 | billion |
| 296 | accord |
| 260 | increase |
| 239 | call |
| 201 | year |
| 198 | change |
| 178 | support |
| 154 | proposal |
| 154 | percent |
| 143 | money |
| 142 | plan |
| 139 | cut |
| 130 | aid |
| 124 | program |
| 122 | people |

Figure 1: The most common nouns preceding the most common prepositions following 'propose', and appearing in the same environment.

object. We sort these verbs by frequency.

7. This sorted list is the list of candidate support verbs for the nominalization.

This method assumes that the verb and the nominalized form of the verb are given. We have experimented with automatically extracting the nominalized form by using the prepositional patterns extracted for the verb in step 5. We extracted 6 megabytes of newspaper articles containing a form of the verb *propose: propose, proposes, proposed, proposing*. Since one use of nominalization is to avoid repetition of the verb form, we suppose that the nominalization of *propose* is likely to appear in the same articles. We extracted the three most common prepositions following a form of *propose* (step 5). We then extracted the nouns appearing in these same articles and which preceded these prepositions. The results[4] appear in figure 1. Since a nominalized form is normally morphologically related to the verb form, almost any morphological comparison method will pick *proposal* from this list.

---

[4] Further experimentation has confirmed these results, but indicate that it may sufficient to simply tag a text, and perform morphological comparison with the most commonly cooccurring nouns in order to extract the nominalized forms of verbs.

| frequency | |
|---|---|
| 167 | reject |
| 127 | hear |
| 114 | make |
| 81 | file |
| ... | |

Figure 2: Most common verbs of which 'appeal' is marked as direct object.

## 4 Experiment with *appeal–appeal*

We have taken for example the case of the verb *appeal* which was interesting since its corresponding deverbal noun shares the same surface form *appeal*. In order to extract corpus evidence, we used a lexical transducer of English that, given the surface word *appeal*, produced all the inflected forms *appeal, appeal's, appealing, appealed, appeals* and *appeals'*.

Using these surface forms as a filter, we scanned 134 Megabytes of tokenized Associated Press newswire stories from the year 1989[5]. As a result of filtering, 6704 sentences (1 Mbyte of text) were extracted. This text was part-of-speech tagged using the Xerox HMM tagger (Cutting et al., 1992). The lexical entries corresponding to *appeal* were tagged with the following tags: as a noun (3910 times), as an active or infinitival verb (1417), as a progressive verb (292), and as a past participle (400).

This tagged text was then parsed by a low-level dependency parser (Grefenstette, 1994)[Chap 3]. From the output of the dependency parser we extracted all the lexically normalized verbs of which *appeal* was tagged as a direct object. The most common of these verbs are shown in Figure 2.

Our speaker's intuition tells us that the support verb for the nominalized use of *appeal* is *make*. But this data does not give us enough information to make this judgement, since concrete versions as a separate entity are not distinguishable from nominalizations of the verb.

In order to separate nominalized uses of the predicate *appeal* from concrete uses, we will refer to the linguistic discussion presented in the introduction that says that nominalizations retain some of the argument/adjunct structure of the verbal predicate. This is verified in the corpus since we find many parallel structures involving *appeal* both as a verb and as a noun, such as:

> Vice President Salvador Laurel said today that an ailing Ferdinand Marcos may not survive the year and **appealed to President Corazon Aquino** to allow her ousted predecessor to die in his homeland.

> Mrs. Marcos made **a public appeal to President Corazon Aquino** to allow Marcos to return to his homeland to die.

Indeed, if we examine a common nominalization transformation, i.e. that of transforming the direct object of a verb into a Norman genitive of the nominalized form, we find a great overlap in the lexical arguments[6].

| | VERB FORM | | NOMINALIZATION |
|---|---|---|---|
| 67 | appeal **decision** | 22 | appeal of **conviction** |
| 61 | appeal **ruling** | 10 | appeal of **ruling** |
| 35 | appeal **conviction** | 9 | appeal of **order** |
| 29 | appeal **verdict** | 6 | appeal of **decision** |
| 23 | appeal case | 4 | appeal of **verdict** |
| 22 | appeal **sentence** | 4 | appeal of **sentence** |
| 21 | appeal **order** | 4 | appeal of plan |
| 7 | appeal judgment | 4 | appeal of inmate |

The parser's output allowed us to extract patterns involving prepositional phrases following noun phrases headed by *appeal* as well as those following verb sequences headed by *appeal*. The most common prepositional phrases found after *appeal* as a verb began with the prepositions[7]: *to* (466 times), *for* (145), *in* (18), *on* (12), *with* (5), etc. The prepositional phrases following *appeal* as a noun are headed by *to* (321 times), *for* (253), *in* (200), *of* (134), *from* (78), *on* (34), etc.

The correspondence between the most frequent prepositions allowed us to consider that the patterns of a noun phrase headed by *appeal* followed by one these prepositional phrases (i.e., begun with *to, for,* and *in*) constituted true nominalizations[8]. There were

---

[5]This corresponds to 20 million words of text.

[6]We decided not to use this type of data in our experiments because matching lexical arguments requires much larger corpora than the ones we had extracted for the other verbs tested.

[7]We ignored prepositional phrases headed by *by* as being probable passivizations, since our parser does not recognize passive patterns involving *by*.

[8]Here we used only part of the corpus evidence that was available. Other patterns of nominalizations of appeal, e.g. Saxon genitives like *the criminal's appeal*, may well exist in the corpus.

|   | frequency |       |
|---|-----------|-------|
|   | 63        | make  |
|   | 16        | have  |
|   | 15        | issue |
|   | ...       |       |

Figure 3: Most common verbs supporting the structure NP PP where 'appeal' heads the NP and where one of {to, for, in} begins the PP.

774 instances of these patterns.

The parser's output further allowed us to extract the verbs for which these nominalizations were considered as the direct objects. 318 of these nominal syntactic patterns including *to, for* and *in* were found. Of these patterns, the main verb supporting the objective nominalizations are shown in Figure 3.

These results suggest that the support verb for the nominalization of *appeal* is *make*.

## 5 Other Predicate Examples and Discussion

When the same filtering technique is applied to subcorpora derived for other nominalization pairs, we obtain the results given in Figure 4. For each verb–noun pair all sentences containing any form of the words were extracted from the AP corpus. The sentences were processed as explained in section 3. For each verbal use, the most frequent prepositional phrases following the verbs were tabulated and the three most frequent prepositions were retained. For example, the most frequent prepositions beginning prepositional phrases following verb uses of the lemma *offer* were *for, in* and *to*. These prepositions were used to select probable nominalizations by extracting noun uses of the predicates that were immediately followed by prepositional phrases headed by one of the three most frequent verbal prepositions. For these extracted noun phrases, when they were found in a direct object position, the main verbs were tabulated which gives the results in Figure 4.

Some of the results in Figure 4 correspond to our naive intuitions of collocational support verbs, such as *make an offer*. For *discussion*, both *have* and *hold* appear equally frequently. But other words show the limitations of this method, we would expect *make a demand* where we find *meet a demand*. In the same subcorpus, although we find *make a demand* 77 times, *meet a demand* is two-and-a-half times more common. Could this be because, in a newswire corpus, *meeting a demand* is more newsworthy than *making* one? If we just look at the cases where *demand* is modified by the indefinite article, which might correspond to the more generic nominalizations one spontaneously creates when generating examples, we find that in the corpus *make a demand* occurs slightly more often than *meet a demand*, ten times vs. six times, but this is too rare to use as a criterion.

In other cases, such as with *proposal* and *assertion* we find *make* and *reject* with almost equal frequency, and though *make* might well be considered a support verb, it is hard to accept *reject* as semantically empty. Though *reject* is more a consequence than an antonym of *make a proposal*, this raises the question, to which we have no answer, of whether support verbs have an equally empty antonym.

A more interesting case is the appearance of *issue* for *order* and *warning* where we would expect *give*. Looking into the corpus evidence, we find *issue a restraining order* 46 times, and *give* any type of order only 16 times. This evidence suggest a limitation of our word-based approach. Multi-word phrases, such as the nominalized phrase *restraining order*, might take a different support verb than the simple unqualified word forms, such as *order*.

## 6 Conclusions

Nominalization is a very productive process. The proper choice of collocational support verbs for nominalizations in English is a difficult task for language learners given the unpredictability of the semantically emptied verb that fulfills the syntactic role. Given a robust parser and large corpus, the simple technique of extracting the most common verbs for which the nominalized form is the direct object is not always sufficient, since completely deverbal concrete noun uses share the same lexical surface form. Comparing argument/adjunct structures involving the verbal uses of the predicate and using the most common of these structures as filters on the surface forms possibly corresponding to nominalizations captures the linguistic fact that nominalizations retain the syntactic structures of their underlying predicate. When these filters are applied, the most common supporting verb in the corpus for the recognized nominalized patterns seems to correspond to native speakers' intuition of the support verb associated with the nominalization.

The experiment described here on a 134

| nominalization | preps | most common main verbs |
|---|---|---|
| offer–offer | for(116), in(100), to(98) | make (116 cases), begin(37), launch(36) |
| discuss–discussion | with(127), in(85), at(54) | have (42), hold(42), begin(9) |
| demand–demand | for(37), in(28), of(22) | meet(58), press(34), increase(22) |
| propose–proposal | in(103), for(77), to(46) | make(28), reject(26), submit(19) |
| order–order | of(91), to(50), in(33) | issue (24), give(8), bring(7) |
| complain–complaint | about(183), of(155), to(91) | receive (20), file(12), have(10) |
| warn–warning | of(140), against(46), in(44) | issue (17), receive(5), make(4) |
| confirm–confirmation | in(30), of(28), to(10) | win (6), recommend(5), have(4) |
| assert–assertion | in(12), at(3), to(2) | make (3), repeat(1), dispute(1) |
| suggest–suggestion | to(60), in(57), of(27) | make(5), reject(5), offer(2) |

Figure 4: Most common verbs supporting the structure found for other nominalization pairs using the syntactic structure filtering mechanism.

megabyte corpus of newspaper text from which was extracted evidence for *appeal* and other predicates shows how this automated procedure can be applied to any verb-nominalization pair given a large corpus and a robust parser. Other work in automated support verb discovery using bilingual dictionaries as a source has been reported in Fontenelle (1993).

It remains to be seen whether these statistical results are more useful to lexicographers than their more traditional tools of key-word-in-context files and T-score measures. Human experiments would be necessary to demonstrate this. Another useful test of the results would be to compare the results given by this technique against machine-readable dictionary-derived data.

In conclusion, the interest of this technique is its general approach to corpus linguistics as one of multiple passes over the same corpus material, using results of previous passes to filter and refine data extracted on subsequent passes. We believe that this approach, coupled with lexical resources and robust parsers, offers much promise for the future of corpus exploitation.